\documentclass[]{article}
\usepackage{proceed2e}
\usepackage{times}
\usepackage{mathtools}

\usepackage{qiangstyle}

\title{Stein Variational Policy Gradient}

\author{Yang Liu\\ UIUC \\ liu301@illinois.edu
 \And Prajit Ramachandran \\ UIUC \\ prajitram@gmail.com \And Qiang Liu \\ Dartmouth \\qiang.liu@dartmouth.edu \And Jian Peng \\ UIUC \\ jianpeng@illinois.edu} 

\usepackage{proceed2e}

\begin{document}

\maketitle

\begin{abstract}
Policy gradient methods have been successfully applied to many complex reinforcement learning problems. However, policy gradient methods suffer from high variance, slow convergence, and inefficient exploration. In this work, we introduce a maximum entropy policy optimization framework which explicitly encourages parameter exploration, and show that this framework can be reduced to a Bayesian inference problem. We then propose a novel Stein variational policy gradient method (SVPG) which combines existing policy gradient methods and a repulsive functional to generate a set of diverse but well-behaved policies. SVPG is robust to initialization and can easily be implemented in a parallel manner. On continuous control problems, we find that implementing SVPG on top of REINFORCE and advantage actor-critic algorithms improves both average return and data efficiency. 
\end{abstract}
\section{Introduction}
Recent advances in policy gradient methods and deep learning have demonstrated their applicability for complex reinforcement learning problems. Instead of deriving a policy from a value function, policy gradient methods directly optimize a parametrized policy with gradients approximated from rollout trajectories. Deep neural networks trained with policy gradient methods have demonstrated impressive performance on continuous control, vision-based navigation and Atari games \citep{schulman2015high, kakade2002natural, schulman2015trust,mnih2016asynchronous}. However, these algorithms are not yet applicable for hard real-world tasks, partially due to  high variance, slow convergence, and insufficient exploration. Because of the non-convexity of neural-network policies, the performance of trained agents is often very sensitive to their initializations. 

In this work, we introduce the Stein variational policy gradient (SVPG) method, a new policy optimization method that leverages a recent Stein variational gradient descent method \citep{liu2016stein} to allow simultaneous exploitation and exploration of multiple policies. Unlike traditional policy optimization which attempts to learn a single policy, we model a distribution of policy parameters, where samples from this distribution will represent strong policies. We first introduce a framework that optimizes this distribution of policy parameters with (relative) entropy regularization. The (relative) entropy term explicitly encourages exploration in the parameter space while also optimizing the expected utility of polices drawn from this distribution. We show that this framework can be reduced to a Bayesian inference problem in which we generate samples of policy parameters from a posterior. We then use Stein variational gradient descent (SVGD) to optimize this distribution. SVGD leverages efficient deterministic dynamics to transport a set of particles to approximate given target posterior distributions. It combines the advantages of MCMC, which does not confine the approximation within a parametric family, and variational inference, which converges fast due to deterministic updates that utilize the gradient. Specifically, in SVPG a policy gradient estimator is applied to improve policy parameter particles while a repulsive functional is used to diversify these particles to enable parameter exploration.

In our experiments, we implementing SVPG on top of existing policy gradient methods, including REINFORCE \citep{williams1992simple} and advantageous actor critic, improves average return on continuous control problems. The SVPG versions enjoy better data efficiency through explicit parameter exploration. SVPG is also more robust to different initializations.  Because of its simplicity and generality, we believe that SVPG provides a generic tool that can be broadly applied to boost the performance of existing policy optimization methods. 

\section{Preliminaries}
We introduce the background of reinforcement learning and discuss different policy gradient estimation methods. 
\subsection{Reinforcement Learning}
Reinforcement learning considers agents which are able to take a sequence of actions in an environment and experience at most one scalar reward per action. The task of the agents is to learn a policy that maximizes cumulative rewards.

Formally, consider an agent operating over time $t\in\{1, \ldots, T\}$. At time $t$ the agent is in an environment state $s_t$ and produces an action $a_t\in A$. The agent will then observe a new state $s_{t+1}$ and receive a scalar reward $r_t \in R$. The set of possible actions $A$ can be discrete or continuous. The goal of reinforcement learning is to find a policy $\pi(a_t|s_t)$ for choosing an action in state $s_t$ to maximize a utility function (or expected return) $$J(\pi) = \mathbf{E}_{s_0,a_0,...}[ \sum_{t=0}^\infty \gamma^t r(s_t,a_t)],$$ where $0\le \gamma \le 1$ is a discount factor; $a_t \sim \pi(a_t|s_t)$ is drawn from the policy; $s_{t+1} \sim P(s_{t+1}|s_t,a_t)$ is generated by the environmental dynamics. In this work, we consider policy $\pi(a_t|s_t)$ to be a stochastic policy which defines the distribution over actions given the current state $s_t$. In the model free setting, we assume that  the environmental dynamics $P(s_{t+1}|s_t,a_t)$ is unknown. The state value function $$V^\pi(s_t)=\mathbf{E}_{a_t,s_{t+1},...}[ \sum_{i=0}^\infty \gamma^i r(s_{t+i},a_{t+i})]$$ is the expected return by policy $\pi$ from state $s_t$. The state-action function $$Q^\pi(s_t,a_t)=\mathbf{E}_{s_{t+1},a_{t+1},...}[ \sum_{i=0}^\infty \gamma^i r(s_{t+i},a_{t+i})]$$ is the expected return by policy $\pi$ after taking action $a_t$ at state $s_t$.
\subsection{Policy Gradient Estimation}
In policy-based reinforcement learning approaches, policy $\pi(a|s;\theta)$ is parameterized by $\theta$ and is iteratively improved by updating $\theta$ to optimize the utility function $J(\theta)$; here by an abuse of notation, we denote $J(\theta) = J(\pi(a|s;\theta)$. There are two main approaches to estimate policy gradient from rollout samples.

\textbf{Finite difference methods.}
It is possible to estimate the gradient  $\nabla_\theta J(\theta)$ using finite difference methods. Instead of computing the finite difference for each individual parameter, rollout efficient methods such as SPSA \citep{spall1998overview}, PEPG \citep{sehnke2010parameter} and evolutionary strategy approximations \citep{hansen2006cma,mannor2003cross} have been proposed. The key idea of such methods is to use a small number of random perturbations to approximate the gradient values of all parameters simultaneously. In particular, the following stochastic finite difference approximation has recently been shown to be highly effective over several complex domains \citep{salimans2017evolution}:
\begin{equation}\label{ES}
\nabla_\theta J(\theta) \approx\frac{1}{m} \sum_{i=1}^m \frac{J(\theta+ h\xi_i)\xi_i}{h},
\end{equation}
where  $\xi_i$ are  drawn from standard Gaussian $\normal(0, I)$. When $h$ is very small, this estimate provides an accurate approximation of $\nabla_\theta J(\theta)$.

\textbf{Likelihood ratio methods.} Policy gradient algorithms, such as the well-known REINFORCE \citep{williams1992simple}, estimate the gradient $\nabla_\theta J(\theta)$ from rollout samples generated by $\pi(a|s;\theta)$ using the likelihood ratio trick. Specifically, REINFORCE uses the following approximator of the policy gradient  
\begin{equation} \label{PG}
\nabla_\theta J(\theta) \approx \sum_{t=0}^\infty \nabla_\theta \log\pi(a_t|s_t;\theta)R_t
\end{equation}
based on a single rollout trajectory, where $R_t=\sum_{i=0}^\infty \gamma^i r(s_{t+i},a_{t+i})$ is the accumulated return from
time step $t$. It was shown that this gradient is an unbiased estimation of $\nabla_\theta J(\theta)$. With a careful choice of baseline function $b(s_t)$, another estimate of policy gradient
\begin{equation}\label{PG_Baseline}
\nabla_\theta J(\theta) \approx  \sum_{t=0}^\infty \nabla_\theta \log\pi(a_t|s_t;\theta)(R_t-b(s_t))
\end{equation}
was shown to be unbiased but with smaller variance. It is common to use the value function $V^\pi(s_t)$ as the baseline. $R_t-V^\pi(s_t)$ is an estimate of advantage function $A^\pi(s_t,a_t) = Q^\pi(s_t,a_t) - V^\pi(s_t)$. 
\begin{equation}\label{PG_AC}
\nabla_\theta J(\theta) \approx  \sum_{t=0}^\infty \nabla_\theta \log\pi(a_t|s_t;\theta)(R_t-V^\pi(s_t))
\end{equation}
This method is also known as advantage actor critic (A2C) \citep{schulman2015high,mnih2016asynchronous}. In practice, multiple rollouts are sometimes used to estimate these policy gradient in a batch.

\section{Stein Variational Policy Gradient}
This section introduces our main framework. 
We start with introducing a maximum entropy policy optimization framework for learning upper level policies in Section~\ref{sec:me}, and then develop our Stein variational policy gradient in Section~\ref{sec:spg}.
\subsection{Maximum Entropy Policy Optimization}\label{sec:me}
Instead of finding a single policy $\theta$, here we consider the policy parameter 
$\theta$ as a random variable and seek a distribution $q(\theta)$ to optimize the expected return. 
We also introduce a \emph{default} parameter  distribution $q_0$ to incorporate prior domain knowledge of parameters or provide the regularization of $\theta$.
We formulate the optimization of $q$ as the following regularized expected utility problem: 
\begin{equation} \label{Opt}
\max_q \bigg\{ \mathbf{E}_{q(\theta)} [J(\theta)] - \alpha \mathbf{D}(q\|q_0) \bigg\}, 
\end{equation}
where $q$ maximizes the expected utility, regularized by a relative entropy or Kullback-Leibler (KL) divergence $\mathbf{D}(q \| q_0)$ with a ``prior distribution'' $q_0$, 
 $$
\mathbf{D}(q\|q_0) = \mathbf{E}_{q(\theta)} [\log q(\theta)-\log q_0(\theta)].
 $$
If we use an uninformative (and improper) prior $q_0(\theta) = const$, the second KL term is simplified to 
the entropy $\mathbf{H}(q) = \mathbf{E}_{q(\theta)} [-\log q(\theta)]$ of $q$. The optimization 
\begin{equation} \label{Opt_ME}
\max_q \bigg\{ \mathbf{E}_{q(\theta)}[J(\theta)] + \alpha \mathbf{H}(q) \bigg\},
\end{equation}
then explicitly encourages exploration in the $\theta$ parameter space according to the principle of maximum entropy. 

By taking the derivative of the objective function in (\ref{Opt}) and setting it to zero, we can show that the optimal distribution of policy parameter $\theta$ of the above optimization is 
\begin{equation} \label{Bayes}
q(\theta) \propto \exp\left (\frac{1}{\alpha}J(\theta)\right) q_0(\theta). 
\end{equation}
This formulation is equivalent to a Bayesian formulation of parameter $\theta$, where $q(\theta)$ can be seen as the ``posterior" distribution; $\exp(J(\theta)/\alpha)$ is the ``likelihood" function; $q_0(\theta)$ is the prior distribution. 
The coefficient $\alpha$ can be viewed as a temperature parameter that controls the strength of exploration in the parameter space. When $\alpha \rightarrow 0$, samples drawn from $q(\theta)$ will be around global optima (or optimum) of the expected return $J(\theta)$. 

A similar idea of entropy regularization has been investigated in the context of bounded relational decision-makers to trade off utility and information costs \citep{wolpert2006information,ortega2011information}. Learning upper level policies are also discussed in \citep{salimans2017evolution}, in which the $q(\theta)$ distribution is often assumed to be follow certain parametric form, such as a Gaussian distribution, whose parameters are optimized to maximize the expected reward (their method does not use entropy regularization and hence is non-Bayesian). Our method does not require a parametric assumption on $q(\theta)$. In many other works, entropy regularization has been proposed directly on policy $\pi$. For example, TRPO \citep{schulman2015trust} enforces a relative entropy constraint to make sure the current policy stays close to a previous policy. Recently \citep{haarnoja2017reinforcement} proposed to leverage the maximum entropy principle on policies and derived a deep energy-based policy optimization method.

\subsection{Stein Variational Gradient Descent}\label{sec:spg}
Traditional Bayesian inference approaches that are used to draw samples from the posterior distribution $q(\theta)$, such as MCMC, may suffer from slow convergence or oscillations due to the high-variance and stochastic nature of estimating $J(\theta)$. Accurate estimation of $J(\theta)$ for a single policy often requires a large number of rollout samples. Instead of utilizing $J(\theta)$, we hope to use the gradient information $\nabla_\theta J(\theta)$ which provides a noisy direction in which to update the policy. For this purpose, here we use the Stein variational gradient descent (SVGD) for Bayesian inference \citep{liu2016stein}.  SVGD is a nonparametric variational inference algorithm that leverages efficient deterministic dynamics to transport a set of particles $\{\theta_i\}_{i=1}^n$ to approximate given target posterior distributions $q(\theta)$. 
SVGD combines the benefits of typical MCMC and variational inference in that it does not confine the approximation within parametric families, unlike traditional variational inference methods, 
and converges faster than MCMC due to the deterministic updates that efficiently leverage gradient information. Thus SVGD has a number of advantages that are critical for policy exploration and optimization.

Briefly, SVGD iteratively updates the ``particles'' $\{\theta_i\}$ via 
$$
\theta_i \gets \theta_i + \epsilon \phi^*(\theta_i), 
$$
where $\epsilon$ is a step size
and $\phi(\cdot)$ is a function in the unit ball of a reproducing kernel Hilbert space (RKHS) $\H = \H_0\times \cdots \H_0$ of $d\times 1$ vector-valued functions, 
chosen to maximumly decrease the KL divergence between the particles and the target distribution in that sense that 
$$
\phi^* \gets \max_{\phi \in \H} \big \{  -\frac{d}{d\epsilon}\mathbf{D}(\rho_{[\epsilon\phi]}  \| q), ~~~~ s.t.~~~~ ||\phi ||_{\H} \leq 1 \}, 
$$
where $\rho_{[\epsilon\phi]}$ denotes the distribution of $\theta' = \theta + \epsilon \phi(\theta)$, and the distribution of $\theta$ is $\rho$.  
\citet{liu2016stein} showed that this functional optimization yields a closed form solution, 
$$
\phi^*(\theta) = \E_{\vartheta\sim\rho} [\nabla \log q(\vartheta)  k(\vartheta, \theta) + \nabla_{\vartheta} k(\vartheta, \theta)], 
$$
where $k(x,x')$ is the positive definite kernel associated with the RKHS $\H_0$. 
By replacing the expectation $\E_{\vartheta\sim\rho}$ with an empirical averaging on the current particles $\{\theta_i\}$, 
we can derive the following Stein variational gradient  
\begin{equation}
\label{equ:svgd}
\begin{split}
& \hat\phi(\theta_i) = \frac{1}{n}\sum_{j=1}^n[\nabla_{\theta_j} \log q(\theta_j)  k(\theta_j, \theta_i) + \nabla_{\theta_j} k(\theta_j, \theta_i)].
\end{split}
\end{equation}
In this update, 
$\hat\phi$ includes two important terms that play different roles. 

\begin{algorithm}[tb] %
\caption{Stein Variational Policy Gradient}  \label{alg:algexample}
\begin{algorithmic}
\STATE \textbf{Input}:  Learning rate $\epsilon$, kernel $k(x,x')$, temperature, 
initial policy particles $\{\theta_i\}$. 
\FOR{iteration $t=0,1,..,T$}
\FOR{particle $i=0,1,..,n$}
\STATE Compute $\nabla_{\theta_i} J(\theta_i)$ e.g., by \eqref{ES}-\eqref{PG_AC}. \\
\ENDFOR \\

\FOR{particle $i=0,1,..,n$}
\STATE
\vspace{-0.5cm} \begin{align*}
\hspace{-0.3cm}  \Delta \theta_i \gets  \frac{1} {n}\sum_{j=1}^n & [ \nabla_{\theta_j} \left(\frac{1}{\alpha} J(\theta_j) + \log q_0(\theta_j)\right)  k(\theta_j, \theta_i) \\ 
& + \nabla_{\theta_j} k(\theta_j, \theta_i)] \\
& \hspace{-1.7cm} \theta_i \gets \theta_i + \epsilon \Delta \theta_i
\end{align*}
\ENDFOR \\
\ENDFOR
\end{algorithmic}
\end{algorithm}

The first term involves the gradient $\nabla_\theta \log q(\theta)$ which drives the policy particles $\theta_i$ towards the high probability regions of $q$ with information sharing across similar particles. The second term $\nabla_{\theta_j} k(\theta_j, \theta_i)$ pushes the particles away from each other, diversifying the policies.  
Note that without the second term, 
all the particles would collapse to the local modes of $\log q$ and the algorithm would reduce to the typical gradient ascent for maximum a posteriori (MAP). 
In particular, if we just use one particle ($n=1$) and choose the kernel with $\nabla_\theta k(\theta,\theta) = 0$, then SVGD is reduced to a single chain of gradient ascent for MAP. 

Applying SVGD to the posterior in \eqref{Bayes}, 
we introduce the Stein Variational Policy Gradient method (SVPG) that 
estimates the policy gradient $\nabla_\theta J(\theta)$ using existing methods, such as those we introduced in \eqref{ES} -\eqref{PG_AC}.  
\begin{equation}
\label{equ:svgd}
\begin{split}
\hat\phi(\theta_i) = \frac{1} {n}\sum_{j=1}^n & [\nabla_{\theta_j} \left(\frac{1}{\alpha} J(\theta_j) + \log q_0(\theta_j)\right)  k(\theta_j, \theta_i) \\ & + \nabla_{\theta_j} k(\theta_j, \theta_i)]
\end{split}
\end{equation}
Note that here the magnitude of $\alpha$ adjusts the relative importance between the policy gradient and the prior term $\nabla_{\theta_j} \left(\frac{1}{\alpha} J(\theta_j) + \log q_0(\theta_j)\right)k(\theta_j, \theta_i)$
and the repulsive term $\nabla_{\theta_j} k(\theta_j, \theta_i)$. A suitable $\alpha$ provides a good trade-off between exploitation and exploration. If $\alpha$ is too large, the Stein gradient would only drive the particles to be consistent with the prior $q_0$.
As $\alpha \to 0$, this algorithm is reduced to running $n$ copies of independent policy gradient algorithms, if $\{\theta_i\}$ are initialized very differently. In practice, we found that with a reasonable $\alpha$, SVPG consistently outperforms the original versions of several policy gradient methods on continuous control tasks.  
A careful annealing scheme of $\alpha$ allows efficient exploration in the beginning of training and later focuses on exploitation towards the end of training.  


\section{Related Work}

Policy gradient techniques have recently shown strong results for deep reinforcement learning. Trust region policy optimization \citep{schulman2015trust} optimizes its policy by establishing a trust region centered at its current policy. Asynchronous advantage actor-critic (A3C) \citep{mnih2016asynchronous} runs multiple agents asynchronously in parallel to train an actor-critic architecture. Numerous works have built on top of A3C \citep{jaderberg2016reinforcement,mirowski2016learning} to solve difficult environments like navigating a 3D maze \citep{beattie2016deepmind}. Recent work has also explored sample efficient policy gradients \citep{gu2016q,wang2016sample}. Notably, all such methods and improvements can be applied to SVPG without any modification.

There has been a wealth of prior work to encourage exploration in reinforcement learning. For example, entropy penalties \citep{williams1991function,mnih2016asynchronous} are used to learn a smoother policy by penalizing the model when the distribution over actions is too sharp. Recent work \citep{nachum2016improving} has explicitly attempted to explore 'under-appreciated' rewards through importance sampling of trajectories over the current policy.

Intrinsic motivation techniques have recently gained popularity in the deep reinforcement learning setting. These techniques \citep{singh2004intrinsically,storck1995reinforcement} attempt to train the agent with rewards beyond those given from the environment. For example, the notion of curiosity can be used to encourage exploration by giving the agent additional rewards for visiting novel states. Novelty can be measured though Bayesian methods \citep{mohamed2015variational,houthooft2016vime}, where information gain is used as a proxy for curiosity. Count-based methods \citep{bellemare2016unifying,ostrovski2017count} produce a reward inversely proportional to the number of times a state has been visited, using a density model to approximate visits in large state spaces. SVPG is compatible with these reward shaping \citep{ng1999policy} procedures since it does not impose any constraints on the parameters and gradients.

Prior work has also explored the use of different agents to collect experience in parallel. \citet{nair2015massively} developed a system called GORILA that deployed multiple actors in parallel to collect experience. Agents differed slightly in parameters due to the asynchronous nature of their training procedure. \citet{mnih2016asynchronous} extended this idea by sampling different exploration $\epsilon$ hyperparameters for the $\epsilon$-greedy policy of different agents. Agents with different sets of parameters was also explored by \citet{osband2016deep} to encourage deep exploration. In practice, all their agents were trained with the same gradient update. In contrast, SVPG uses different updates for each agent and explicitly encourages diversity through the kernel gradient term.

\begin{figure*}
  \includegraphics[width=0.99\textwidth]{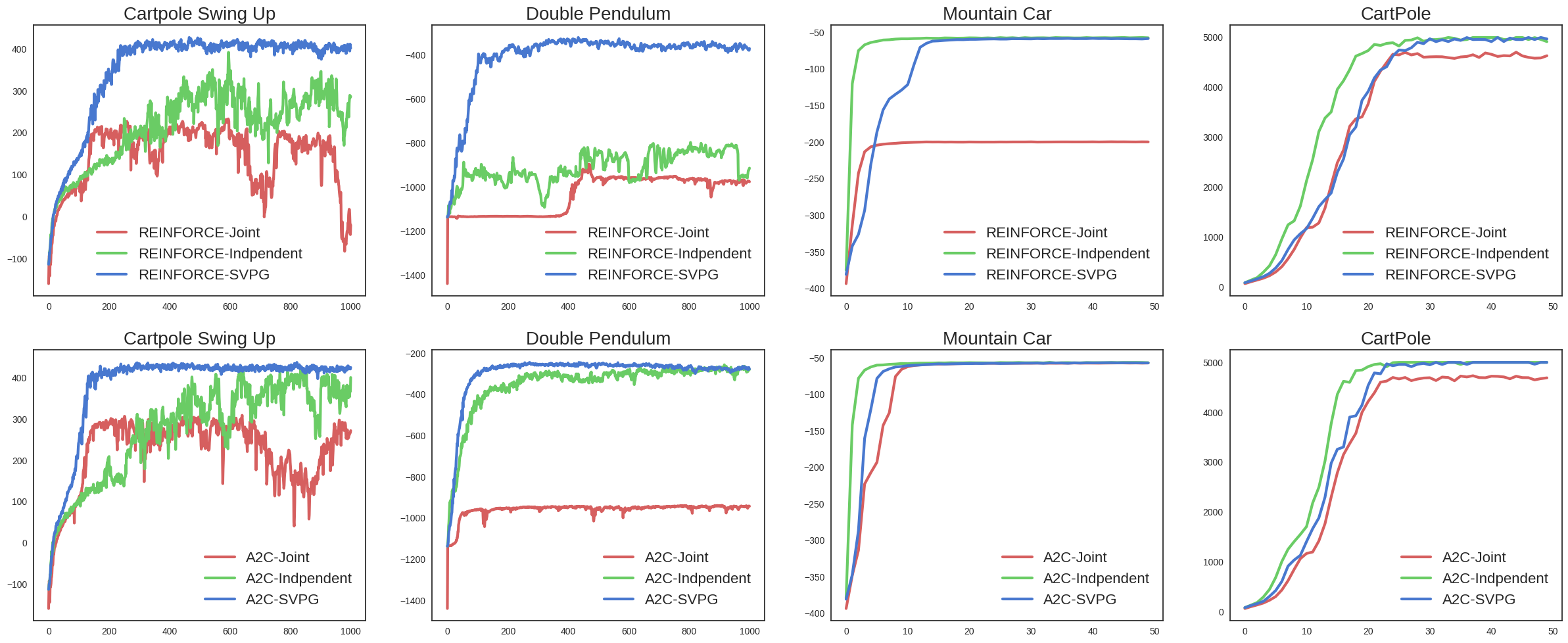}
    \caption{\textbf{Learning curves by SVPG and two baseline versions.} The x-axis denotes the training iteration while the y-axis denotes the average return achieved by the policy. Since all three algorithms use the same number of samples per iteration, the x-axis is also proportional to the total number of samples used in training.}
    \label{results_curves}
\end{figure*}

\begin{table*}
\centering
\caption{Best test return and the number of episodes required to reach within 5\% of the maximum return.}
\label{table:performance}
\begin{tabular}{|l|ll|ll|ll|}
\hline
  A2C             & Joint & episodes & Independent & episodes & SVPG & episodes   \\
\hline
Cartpole Swing Up & 308.71    & 189      & 419.62          & 474      & \textbf{436.84}   & \textbf{171}        \\
Double Pendulum   & -938.73   & 46       & -256.64         & 638      & \textbf{-244.85}  & \textbf{199}        \\
\hline
\hline
REINFORCE         & Joint & episodes & Independent & episodes & SVPG & episodes   \\
\hline
Cartpole Swing Up &  232.96	   &	253	 &   391.46        &   594    &  \textbf{426.69}  &    \textbf{238}    \\
Double Pendulum   &  -892.31	& 446 &  -797.45       &   443    & \textbf{-319.66}  &   \textbf{327}      \\
\hline
\end{tabular}
\end{table*}

\begin{figure*}
\begin{center}
\includegraphics[width=1.0\textwidth]{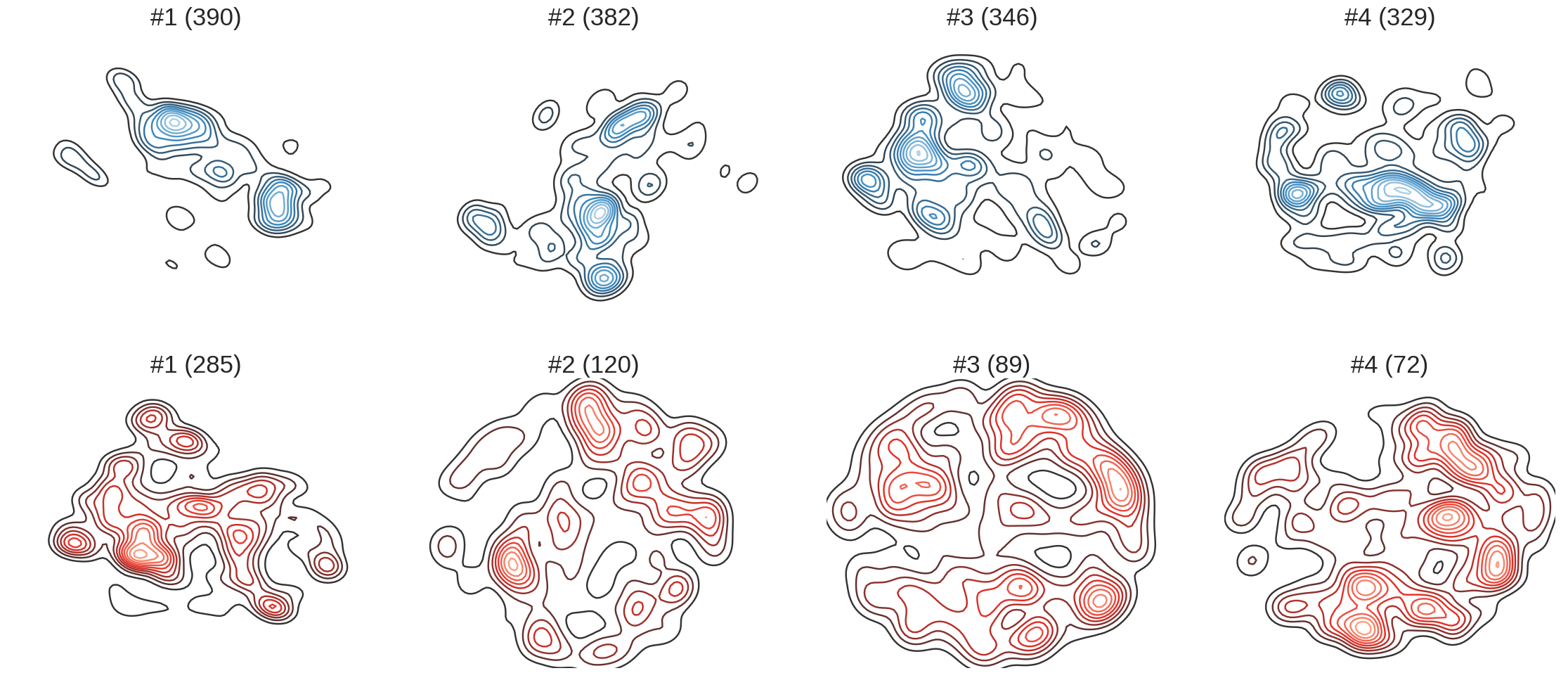}
    \caption{\textbf{State visitation density by REINFORCE-SVPG/Independent algorithms.} The he state visitation landscapes of the best four policies learned by SVPG (first row) and Independent agents (second row). The value in the parenthesis is the average return of a policy. All states are projected from a 4D space into a 2D space by t-SNE \citep{maaten2008visualizing}.}
\label{fig:trajectoris}
\end{center}
\end{figure*}

\section{Experiments}
In this section, we design experiments to 1) compare the SVPG versions of policy gradient algorithms with their original versions in terms of convergence and data efficiency, 2) analyze the characteristics of policies simultaneously learned by SVPG, 3) investigate the trade-off between exploration and exploitation in SVPG. 

For comparisons, we implement the SVPG and the original versions of two popular policy gradient algorithms: REINFORCE (Eq (\ref{PG})) and the advantage actor critic (A2C) (Eq (\ref{PG_AC})).
In the A2C implementations, we use the generalized advantage estimator (GAE) \citep{schulman2015high} for critic, which is implemented in the rllab toolkit \citep{schulman2015high,duan2016benchmarking}. Only the policy parameters are updated by SVPG, while the critics are updated with the their normal TD-error gradient. After training, the agent that gives the best average return is selected. 

We use the flat improper prior $\log q_0(\theta) = 1$ in all our experiments.
For SVGD, we follow \cite{liu2016stein} to use the Gaussian RBF kernel $k(\vartheta,\theta)=\exp(-||\theta-\vartheta||_2^2/h)$ with the bandwidth $h$ chosen to be $med^2/\log(n+1)$ where $med$ denotes the median of pairwise distances between the particles $\{\theta_i\}$.  This simple heuristic allows the bandwidth to adaptively change as the particles move, ensuring that there is always a significant number of particles that interact with each other. 

Since SVPG is a multi-agent method that simultaneously trains a number of policies, we design two implementations of original REINFORCE and A2C for fairness when comparing data efficiency:
\begin{itemize}
\item \textbf{REINFORCE/A2C-Joint}: We train a single agent with the same amount of data as used by multiple agents in SVPG. Assume that we use $m$ transition samples for each of the $n$ agents by SVPG during each iteration, we use in total $nm$ samples per iteration to train a single agent here. In this way, we can also view this joint version as a parallel policy gradient implementation with $n$ threads, each accumulating the gradient information in parallel. This joint variation enjoys better gradient estimation, but SVPG's multiple agents will provide more exploration. This experiment tries to determine whether the repulsive functional in SVPG will encourage exploration that lead to better policies.

\item \textbf{REINFORCE/A2C-Independent}: We train multiple agents independently with no communication among them. So for each of the $n$ agents, we use $m$ samples to calculate the gradient. After training, the agent that gives best average return is selected. This experiment tries to determine the importance of information sharing between different agents through gradients.
\end{itemize}



\subsection{Experimental Setting}
All the experiments are constructed with the OpenAI rllab development toolkit \citep{duan2016benchmarking}. Specifically, we benchmark on four classical continuous control tasks: Cartpole Swing-Up, Double Pendulumn, Cartpole, and MountainCar. The maximal length of trajectories is set to $500$. Details of these environments can be found in \citep{duan2016benchmarking} and on  the GitHub website\footnote{https://github.com/openai/rllab}.  

Here we describe the default setting of the policy gradient algorithms we test on these control tasks. We used a Gaussian policy with a diagonal covariance for all these tasks. The mean is parameterized by a neural networks with three hidden layers (100-50-25 hidden units) and $\tanh$ as the activation function. The log standard deviation is parameterized by a vector, as suggested in \citep{duan2016benchmarking,schulman2015trust}. The SVPG and Independent algorithms used $m=10000$ samples for policy gradient estimation for each of the $n=16$ agents. We found that the choice of $m$ is quite robust as long as it is not too small. 
For the Joint version, the number of samples used in each iteration is $nm=160,000$. For A2C, we set $\lambda=1$ for the generalized advantage estimation as suggested in  \citep{duan2016benchmarking}. In the SVPG algorithms, we set $\alpha=10$, which seems to find a good trade-off between exploration and exploitation in these tasks. We also investigate this hyperparameter in a later experiment. We used ADAM \citep{kingma2014adam} to adjust the step-size of gradient descent in all algorithms. For the two easy tasks, Mountain Car and Cartpole, all agents are trained for 50 iterations with 5 re-runs. For the two complex tasks, Cartpole Swing-Up and Double Pendulum, we train all agents up to $1000$ iterations with 5 re-runs with different random seeds.

\subsection{Convergence and Data Efficiency}
The learning curves of algorithms are shown in Figure \ref{results_curves}. On the two easy tasks, Mountain Car and Cartpole, almost all algorithms can successfully solve them in around 20 iterations (except of REINFORCE-Joint on Mountain Car). On the two more challenging tasks, the differences between algorithms are substantial. On Cartpole Swing-Up, REINFORCE-SVPG and A2C-SVPG converges substantially faster than the corresponding baselines. A2C-Independent is able to achieve the desirable performance after more than 800 iterations, while all other baselines cannot converge after 1,000 iterations. On Double Pendulum, We also performed the same training with a different batch size $m=5000$ and found similar results. The variances of the SVPG algorithms are close to those of the Independent versions, but much smaller than the Joint versions, probably due to the robustness of multiple agents. These results demonstrate that SVPG is able to improve the training convergence of policy gradient algorithms.


\subsection{SVPG learns strong, diverse policies}
We further wanted to investigate the quality and diversity of the policies learned by SVPG. 

First, we computed the best test returns of the policies learned by all algorithms and counted how many episodes each policy needed to achieve the 95\% of its best return. Results are shown in Table \ref{table:performance}. SVPG versions clearly outperforms policies learned by Joint and Independent versions, in terms of both best test returns and the episodes required to achieve 95\% of the return. 

\begin{center}
\begin{figure}[htbp]
    \includegraphics[width=0.50\textwidth]{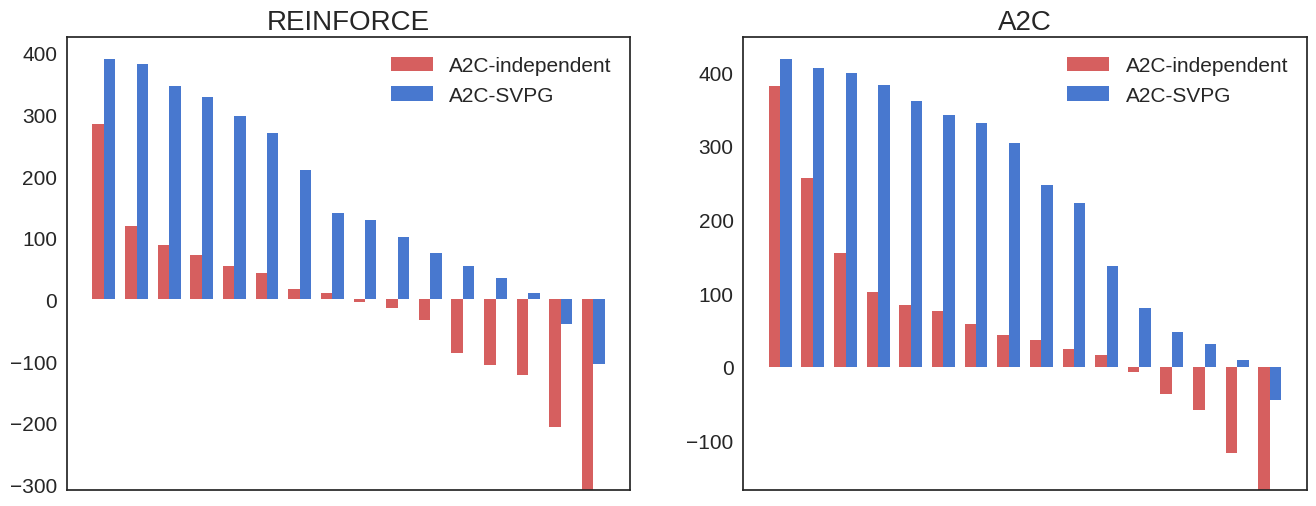}
    \caption{Average return comparison for all agents learned by the SVPG and Independent algorithms.}
    \label{fig:score_per_agent}
\end{figure}
\end{center}

Second, we compare the quality of all policies learned by SVPG and Independent versions. It is worth noting that in  Independent versions, parameter exploration is only done through random initialization, while in SVPG, parameter exploration is explicitly done through repulsion. Thus it is possible that SVPG can better explore the parameter landscape and find multiple good local optima or solutions. We  compute average returns of all policies with 50,000 test transitions. The results are then averaged across 5 re-runs with different training random seeds. The sorted policies are shown in Figure \ref{fig:score_per_agent}. SVPG has found many good policies, while only the very top policies learned by Independent algorithms can achieve satisfactory performance. 

Third, we visualize the diversity of the policies learned by SVPG algorithms. Since SVPG learns policies simultaneously, it is possible that all the top policies are very similar. It is worth noting that good policies cannot be starkly different from each other. To illustrate the diversity of the learned policies in SVPG approach, we generated the state-visitation landscape of individual policies. We chose the best four policies learned by REINFORCE-SVPG and the best four by REINFORCE-Independent and randomly generate $100$ test trajectories for each policy. Then all states are projected into 2D space with the t-SNE algorithm \citep{maaten2008visualizing,van2014accelerating,Ulyanov2016} for visualization. Then we plotted the density contours of state-visitation frequencies for each policy in Figure \ref{fig:trajectoris}. Note that the 2D projection may introduce artifacts in the visualization.

Notwithstanding artifacts, it seems that the state visitation landscapes of the policies learned by SVPG are different from each other. Each policy covers certain regions in the projected state space with high visitation frequencies. Three policies learned by the Independent algorithm (\#2-\#4) are quite similar to each other, but it average return of these policies are far from the best average return. It is worth noting that the best policy by the Independent algorithm looks similar to SVPG policy \#3, and it is possible that they are close to the same local optimum.  


These observations indicate that SVPG can learn robust and diverse policies thanks to the balance between information sharing and repulsive forces. 



\begin{center}
\begin{figure}[htbp]
  \includegraphics[width=0.50\textwidth]{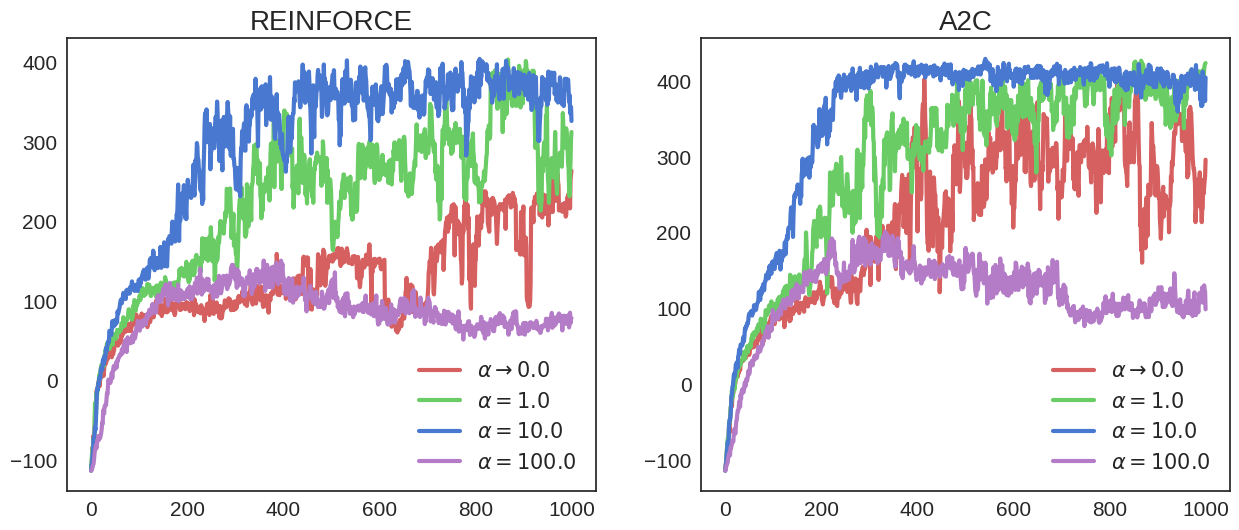}
    \caption{The influence of the temperature hyperparameter in REINFORCE/A2C-SVPG.}
    \label{fig:hyper_temp}
\end{figure}
\end{center}

\subsection{Exploration and Exploitation}
The above results demonstrate that SVPG can better explore in the parameter space than the original policy gradient methods. In SVPG, the temperature hyperparameter $\alpha$ controls the trade-off between exploitation, which is driven by policy gradient, and exploration, which is driven by the repulsion between parameters of different policies. Lower temperatures make the algorithm focus on exploitation, as the first term with the policy gradient becomes much more important than the second repulsive term. Higher temperature will drive the policies to be more diverse. That is,
$$
\sum_{j=1}^n [\underbrace{\nabla_{\theta_j} \frac{1}{\alpha} J(\theta_j) k(\theta_j, \theta_i) }_{
\emph exploitation} ~+~ \underbrace{\nabla_{\theta_j} k(\theta_j, \theta_i)}_{\emph exploration}]. 
$$
Figure \ref{fig:hyper_temp} shows an ablation over different values of the temperature hyperparameter $\alpha$ with batch size $m=5,000$.   

When the temperature is very high (e.g. $\alpha=100$), exploration dominates exploitation. Policies are repelled too much from each other, which prevents them from converging to a good set of policies. When the temperature is too low ($\alpha \rightarrow 0$), exploitation dominates exploration. The algorithm is then similar to training a set of independent policies. As training proceeding, the Euclidean distances between parameters of policies become very large. As a result, both information sharing through kernel and inter-agent exploration are no longer effective. An intermediate  $\alpha = 10$ balances between exploration and exploitation, and outperforms other temperatures for both actor-critic and REINFORCE. We believe that a good annealing scheme of $\alpha$ enables efficient exploration in the beginning of policy training and while focusing on exploitation in the later stage for policy tuning.




\section{Conclusion}
In this work, we first introduced a maximum entropy policy optimization framework. By modeling a distribution of policy parameters with the (relative) entropy regularization, this framework explicitly encourages exploration in the parameter space while also optimizing the expected utility of polices generated from this distribution. We showed that this framework can be reduced to a Bayesian inference problem and then propose the 
Stein variational policy gradient to allows simultaneous exploitation and exploration of multiple policies. In our experiments, we evaluated the SVPG versions of REINFORCE and A2C on several continuous control tasks and found that they greatly improved the performance of the original policy gradient methods. We also performed extensive analysis of SVPG to demonstrate its robustness and the ability to generate diverse policies. Due to its simplicity, we expect that SVPG provides a generic approach to improve existing policy gradient methods. The parallel nature of SVPG also makes it attractive to implement in distributed systems. 

There are a lot of potential future directions based on this work. The simplest extension is to measure the empirical performance of SVPG in other RL domains. Furthermore, the choices of kernel in SVPG may be critical for high dimensional RL problems, and future work should explore the impact of the choice of kernel. For example, it is possible to design a layer-wise kernel for neural network policies. Another direction is to further study the trade-off between exploration and exploitation in SVPG. Currently, we simply choose a fixed $\alpha$ to control the ratio between the policy gradient and the repulsive terms. Smarter annealing schemes of $\alpha$ may lead to a better and adaptive trade-off. Finally, it would be very interesting to extend SVPG for other policy optimization methods such as natural policy gradient and trust region policy optimization.

\bibliographystyle{apalike} 
\bibliography{RL}

\end{document}